%% file: Main.tex
\documentclass[preprint]{elsarticle}
\usepackage{xpatch}
\usepackage{array}
\newcolumntype{P}[1]{>{\centering\arraybackslash}m{#1}}
\usepackage[table]{xcolor}

\usepackage{booktabs}
\usepackage{tabularx}
\usepackage{adjustbox}
\usepackage{multirow}
\usepackage{amsmath,amssymb,amsfonts}
\usepackage{algorithm}
\usepackage{algpseudocode}
\usepackage{graphicx}
\usepackage{textcomp}
\usepackage[flushleft]{threeparttable}
\usepackage{enumitem}
\usepackage[bottom]{footmisc}
\usepackage{longtable}
\usepackage{caption}
\usepackage{subcaption}
\usepackage{afterpage}
\usepackage[normalem]{ulem}
\usepackage{url}
\usepackage[hidelinks]{hyperref}
\usepackage{lineno,hyperref}
\usepackage{float}

\xpatchcmd{\State}{\algorithmicend\ \algorithmicfor}{\algorithmicend}{}{}

\biboptions{sort&compress}

\begin{document}

\begin{frontmatter}

\title{LLM-ADAM: A Generalizable LLM Agent Framework for Pre-Print Anomaly Detection in Additive Manufacturing}

\author[UIUCMechSE,UIUCCSL]{Ahmadreza Eslaminia}

\author[UMich]{Chuhan Cai}

\author[UMich]{Cameron Smith}

\author[UMich]{Ruo-Syuan Mei}

\author[UIUCMechSE]{Shichen Li}

\author[Rut]{Rajiv Malhotra}

\author[UIUCCSL]{Klara Nahrstedt}

\author[UMich,UIUCMechSE]{Chenhui Shao\corref{mycorrespondingauthor}}
\ead{chshao@umich.edu}

\address[UIUCMechSE]{Department of Mechanical Science and Engineering, University of Illinois at Urbana-Champaign, Urbana, IL 61801, USA}
\address[UIUCCSL]{Coordinated Science Laboratory, University of Illinois at Urbana-Champaign, Urbana, IL 61801, USA}

\address[Rut]{Department of Mechanical and Aerospace Engineering, Rutgers University, Piscataway, NJ 08854, USA}

\address[UMich]{Department of Mechanical Engineering, University of Michigan, Ann Arbor, MI 48109, USA}

\cortext[mycorrespondingauthor]{Corresponding author}

\begin{abstract}
Additive manufacturing (AM) continues to transform modern manufacturing by enabling flexible, on-demand production of complex geometries across diverse industries, from aerospace to healthcare. Fused filament fabrication (FFF) has extended AM to laboratories, classrooms, and small production environments, but this accessibility shifts process-planning responsibility to users who may lack manufacturing expertise. A syntactically valid slicer profile can still encode thermally or geometrically harmful settings, and subtle G-code edits can alter extrusion, cooling, or adhesion before a print begins. Pre-print screening from G-code can therefore catch accidental or adversarial machine-program errors before material or machine time is wasted, or part integrity is compromised. This paper proposes LLM-ADAM, where LLM stands for large language model and ADAM for anomaly detection in additive manufacturing, as a generalizable LLM agent framework for pre-print anomaly detection. The framework decomposes the task into three roles: Extractor--LLM maps a G-code file to a structured process-parameter schema; Reference--LLM converts printer and material documentation into aligned operating ranges; and Judge--LLM interprets a deterministic deviation table and selected G-code evidence to decide whether a part is non-defective or belongs to an anomaly class. The decomposition is intended to be model- and machine-agnostic, with specific printers, materials, and proprietary LLM backbones treated as interchangeable test conditions rather than assumptions of the method. We evaluate the framework on an FFF G-code corpus spanning two desktop printer families, two materials, and five classes including non-defective, under-extrusion, over-extrusion, warping, and stringing. A $3\times3$ Reference--LLM$\times$Judge--LLM screening study is first used for model selection. The selected configurations are then evaluated on the full $N=200$ corpus, where the best framework configuration reaches $87.5\%$ accuracy, compared with $59.5\%$ for the strongest engineered single-LLM baseline. The results show that structured decomposition, rather than backbone strength alone, is the dominant source of improvement, with defect classes identified at or near ceiling for leading configurations while residual errors concentrate on conservative false alarms for non-defective samples.
\end{abstract}

\begin{keyword}
Large language model \sep LLM agent \sep Fused filament fabrication \sep Additive manufacturing \sep Anomaly detection \sep G-code analysis \sep Quality control
\end{keyword}

\end{frontmatter}


\input{sections/introduction}
\input{sections/related_work}
\input{sections/problem_formulation}
\input{sections/methodology}
\input{sections/implementation}
\input{sections/case_studies}
\input{sections/discussion}
\input{sections/conclusion}
\input{sections/back_matter}

\end{document}

%% file: sections/introduction.tex
\section{Introduction}\label{sec:introduction}

Additive manufacturing (AM) increasingly supports distributed, flexible, and on-demand production~\cite{abdulhameed2019additive,mehta2022federated}. It has found broad application across aerospace \cite{blakey2021metal,pant2023applications}, biomedical~\cite{kumar2021role,conway2024geometry,sun2025emerging}, automotive~\cite{leal2017additive,salifu2022recent}, and consumer product sectors~\cite{mcgregor2022using,islam2024additive,mehta2024federated}, where its ability to produce complex geometries on demand reduces tooling cost and lead time. Within AM, fused filament fabrication (FFF) has extended this capability to laboratories, makerspaces, and small manufacturers, but the same accessibility that makes FFF tractable for non-expert users also introduces process-planning risk~\cite{singh2020current,cleeman2025scalable}. A slicer profile that appears syntactically valid can encode thermally or geometrically harmful settings; a subtle edit to the resulting G-code can alter extrusion, cooling, or adhesion without changing the visible geometry of the part. Pre-print anomaly detection from G-code addresses this gap directly: it screens the machine program before fabrication begins, catching accidental misconfiguration and adversarial perturbations before a print consumes material, time, or machine capacity.

FFF quality is governed by interacting process variables including extrusion flow, line width, layer height, nozzle and bed temperature, retraction, travel speed, cooling policy, and bed-adhesion strategy, whose effects are not independent~\cite{dey2019systematic,baechle2022failures,kantaros2021employing,cleeman2025operational}. A parameter that is within tolerance in one material or printer profile can become defect-inducing under a different thermal envelope or fan policy. Conventional quality monitoring approaches operate during or after fabrication using images, sensors, or operator feedback, and do not address whether a machine program is safe before it reaches the printer. G-code is a textual representation of toolpath and process instructions~\cite{rivet2023mechanical,ochoa2025digital}, and is therefore a natural substrate for language-model reasoning, provided that reasoning is grounded in the numerical and procedural structure of the manufacturing task rather than treated as free-form text.

Large language models (LLMs) have been explored for AM troubleshooting, process planning, G-code manipulation, and contextual question answering~\cite{sriwastwa2023generative,badini2023chatgpt,jignasu2023towards,chandrasekhar2024amgpt,jadhav2024llm}. These studies demonstrated that current models carry useful technical and code-understanding priors. However, a monolithic prompt that asks one model to read a long G-code file and produce a defect label places several heterogeneous tasks inside one inference step. The model must extract process parameters from repetitive machine code, identify the correct reference envelope for the printer and material, perform numerical comparisons, and decide whether deviations are benign or defect-inducing. A single LLM is unlikely to be equally strong at all of these subtasks, so the solution should be viewed not only as a prompt-engineering issue but also as a systems-design challenge.

This paper proposes LLM-ADAM, where ADAM stands for anomaly detection in additive manufacturing. The framework addresses pre-print G-code screening in FFF through a central hypothesis: LLMs are more reliable when assigned role-specific, schema-constrained subtasks and when numerical comparisons are externalized into deterministic intermediate artifacts rather than delegated to token-level arithmetic. The framework decomposes inference into three stages: Extractor--LLM maps G-code into a fixed process-parameter schema; Reference--LLM maps printer and material documentation into aligned operating ranges; and Judge--LLM evaluates extracted parameters, reference ranges, deviation magnitudes, and selected G-code evidence to produce an anomaly label. The method is not tied to a particular printer, material, anomaly set, or LLM provider. Prusa MK4S (MK4S) and Bambu Lab P1S (BMP1) printers, PLA and ABS materials, and three closed-source LLM providers serve as a controlled testbed for studying the decomposition.

The paper makes three contributions. First, it formulates FFF pre-print anomaly detection as automated reasoning over G-code, documentation-derived reference ranges, and deterministic deviation tables. Second, it introduces a stage-separable LLM agent architecture that allows different model backbones to be assigned to extraction, reference construction, and judgment. Third, it provides an empirical evaluation on a stratified multi-printer, multi-material dataset. A screening study is used to select the strongest framework configurations, and the selected configurations are then evaluated on the full $N=200$ corpus, where the best framework configuration reaches $87.5\%$ accuracy against a 59.5\% single-LLM baseline.

The remainder of this paper is organized as follows. Section~\ref{sec:related} reviews related work. Section~\ref{sec:problem} describes the task and dataset. Section~\ref{sec:method} presents the LLM-ADAM framework. Section~\ref{sec:experiment} describes the experimental protocol. Section~\ref{sec:case} reports the screening and full-corpus results. Section~\ref{sec:discussion} discusses interpretation and limitations, and Section~\ref{sec:conclusion} concludes the paper.

%% file: sections/related_work.tex
\section{Related Work}\label{sec:related}

\subsection{LLMs in additive manufacturing}

LLMs are increasingly studied as interfaces to technical manufacturing knowledge because they can operate over natural language, code, documentation, and process descriptions. In AM, early work has examined whether general-purpose LLMs can answer user questions, support troubleshooting, and assist parameter selection for 3D printing~\cite{sriwastwa2023generative,badini2023chatgpt,chandrasekhar2024amgpt}. Other studies have evaluated LLMs on G-code transformation and comprehension tasks, including translation, rotation, debugging, and toolpath manipulation \cite{jignasu2023towards}. LLMs have also been studied as components of monitoring and control loops~\cite{jadhav2024llm}. These studies demonstrate that LLMs can access useful AM knowledge and reason over manufacturing text, but most treat the model as a single black-box decision maker. For safety-relevant G-code screening, such an interface is insufficient because the model must parse a long machine program, identify process variables, know the valid operating envelope, compare numbers, and decide whether deviations are meaningful. The present work addresses this limitation by separating those functions into role-specific stages and by introducing deterministic intermediate representations~\cite{eslaminia2025fdmbench}.

\subsection{LLM agents}

LLM research has moved from single-prompt answering toward agentic and tool-augmented systems in which a language model interacts with retrieval modules, external tools, structured schemas, and intermediate reasoning artifacts~\cite{lewis2020retrieval,yu2026qosqoe,yao2023react}. This shift is especially relevant for engineering applications. Engineering decisions often require traceable evidence and numerical consistency rather than only plausible prose. Structured-output prompting can improve reliability when the desired answer is a machine-readable object, while deterministic computation can handle comparisons and transformations that should not be delegated to token-level arithmetic~\cite{ni2024l2ceval}. The proposed LLM-ADAM framework follows this automated-agent principle where LLMs perform flexible interpretation of text programs and manuals, while deterministic code constructs the comparison object that the final judge consumes.

In manufacturing, agentic LLM systems remain comparatively underdeveloped. Retrieval-augmented systems can support contextual AM queries, and closed-loop systems can use LLMs as high-level controllers, but pre-print G-code anomaly detection requires a different design~\cite{khanghah2025multimodal,khanghah2026zeroshot}. The system must produce a decision that can be traced to extracted process parameters, reference bands, and the original machine program. This motivates a decomposition in which each stage has a narrow contract and the intermediate artifacts are inspectable. The contribution of this paper is an inference architecture that makes existing backbones more usable and comparable at each step of an auditable manufacturing-screening task.

\subsection{Anomaly detection via G-code inspection}

Quality monitoring and anomaly detection approaches in manufacturing most commonly operate on physical signals collected during or after fabrication (e.g., images, thermal data, acoustic emissions, vibration, 3D point clouds), across domains including AM \cite{mcgregor2021analyzing,yang2022hierarchical,cai2023review,chen2024situ}, machining \cite{shao2017improving,li2025method,xu2025self,li2026smart}, and welding \cite{tian2023weldmon,shao2013feature,eslaminia2024fdgcm, geng2025machine,eslaminia2026adaptive,tian2025machinestethoscope}. 
These approaches are essential for process monitoring, but they are complementary to pre-print screening. A G-code-level method examines the instruction sequence before the machine executes it, which is useful when the goal is to prevent avoidable failures, reduce waste for novice users, or detect suspicious program edits. Prior work has shown that subtle malicious G-code modifications can be difficult to identify with conventional tools~\cite{beckwith2021needle}. LLMs offer a promising interface because they can read both code-like programs and documentation; however, the evidence must be organized so that the model reasons over manufacturing-relevant structure rather than over an undifferentiated token stream. LLM-ADAM provides this organization through a three-stage decomposition and a deterministic deviation representation.

%% file: sections/problem_formulation.tex
\section{Problem Formulation and Dataset}\label{sec:problem}

\subsection{Task definition}

This paper studies pre-print anomaly detection from the G-code program that will be sent to an FFF printer. Because the goal is to detect potential defects before fabrication, the inference setting intentionally excludes images, sensor traces, and operator annotations. Given a G-code file $g$, a printer identifier $p$, and a material identifier $m$, the system predicts a label $\hat{y}$ from five mutually exclusive classes,
\[
\mathcal{Y}=\{\texttt{ND},\texttt{UE},\texttt{OE},\texttt{WP},\texttt{ST}\},
\]
corresponding to non-defective (ND), under-extrusion (UE), over-extrusion (OE), warping (WP), and stringing (ST). The system may also return a confidence score, a rationale, and citations to evidence in the G-code, but the primary quantitative target is the class label. The presence of $p$ and $m$ does not make the task printer-specific; rather, it reflects the fact that a physically meaningful judgment requires comparing the program against the operating envelope of the intended machine and material. A general screening method should therefore support different printer and material combinations through interchangeable reference knowledge.

\subsection{Dataset}\label{sec:dataset}

The experimental corpus is a multi-printer, multi-material FFF dataset professionally designed to separate defect mechanisms while retaining realistic slicer and machine-program structure. Table~\ref{tab:dataset_summary} summarizes the factorial design. The full corpus contains $200$ printed parts, spanning two commercial desktop printers, two materials, five anomaly classes, and ten calibrated parameter variations per cell. 

\begin{table}[t]
\centering
\small
\setlength{\tabcolsep}{6pt}
\caption{Factorial structure of the printed corpus and the extraction benchmark subset. The full corpus size follows from multiplying the levels of printers, materials, anomaly classes, and parameter-variation replicates. The extraction benchmark holds one G-code per $(\text{printer},\text{material},\text{class})$ cell.}
\label{tab:dataset_summary}
\begin{tabularx}{\textwidth}{@{}l X c@{}}
\toprule
Controlled factor & Specification & Levels \\
\midrule
Printers & Prusa MK4S; Bambu Lab P1S  & 2 \\
Materials & PLA; ABS & 2 \\
Anomaly classes & \texttt{ND}, \texttt{UE}, \texttt{OE}, \texttt{WP}, \texttt{ST}  & 5 \\
Parameter variations & Ten settings per $(\text{printer},\text{material},\text{class})$ cell & 10 \\
\midrule
Total printed parts & Full factorial corpus (stratified) & $2{\times}2{\times}5{\times}10=200$ \\

\bottomrule
\end{tabularx}
\end{table}

Class induction was controlled by perturbing parameter families associated with each defect mechanism. UE and OE samples emphasize deposition-volume changes, ST samples emphasize thermal, retraction, and travel behavior, and WP samples emphasize adhesion, bed temperature, and cooling behavior. The corpus was intentionally designed to include multiple parameter combinations that can lead to the same visible anomaly rather than only scaling one parameter by different amounts. It also includes edge cases in which a normally strong indicator, such as elevated extrusion flow for OE, is combined with other parameters in a way that can produce UE or an ND part. Similarly, some ND samples include combinations of settings outside nominal ranges that nevertheless produced acceptable parts. These cases make the benchmark more challenging for LLMs because a classifier must reason over interacting parameters rather than rely on a single cue. Each printed part was visually inspected against its corresponding baseline and assigned exactly one label, so the benchmark remains suitable for supervised single-label classification rather than multi-label diagnosis.

Two subsets are used before the final full-corpus experiment. The $N=20$ extraction benchmark contains one G-code per printer--material--class cell and is used to evaluate parameter extraction against manually recorded ground truth. The $N=40$ end-to-end screening set contains two G-codes per cell and is used to compare Reference--LLM and Judge--LLM combinations. The full $N=200$ corpus is the final evaluation set for the selected multi-stage candidates and the selected single-LLM baseline, as described in Section~\ref{sec:planned_full_corpus}.

%% file: sections/methodology.tex
\section{Methodology}\label{sec:method}

This section describes the proposed automated LLM agent framework. The design goal is to transform a long machine program, such as G-code, and the relevant reference documents into a sequence of auditable intermediate representations before making a final defect judgment. Rather than relying on one model to perform parameter extraction, reference retrieval, numerical comparison, and classification simultaneously, the framework assigns these responsibilities to separate roles connected by deterministic computation.

\subsection{Framework overview}\label{sec:framework}

\begin{figure}[h]
\centering
\makebox[\textwidth][c]{\includegraphics[width=1.05\textwidth]{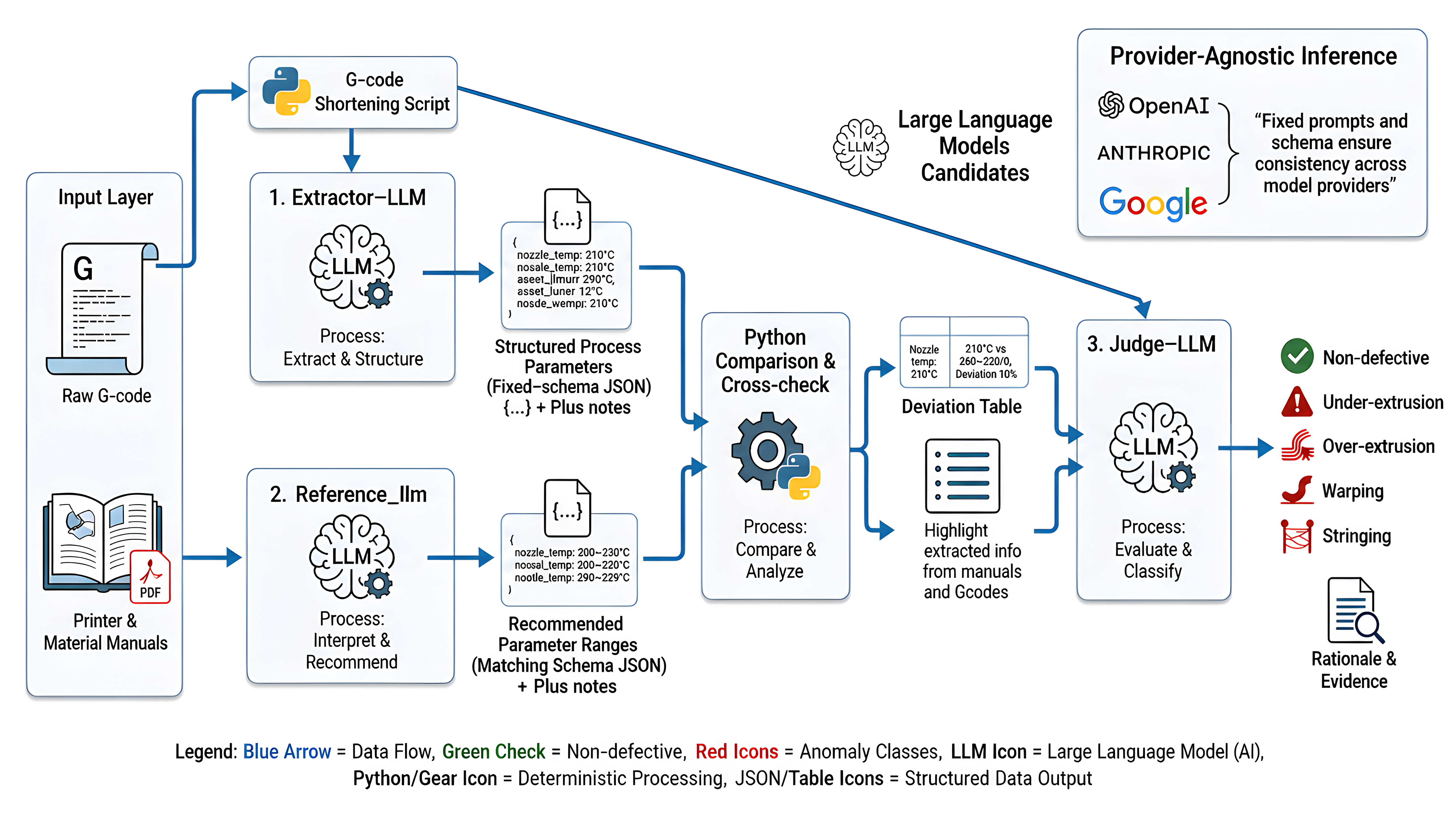}}
\caption{Overview of the proposed LLM agent framework. Extractor--LLM maps G-code into structured process parameters. Reference--LLM maps printer and material documentation into aligned operating ranges. A deterministic comparison layer constructs a deviation table. Judge--LLM consumes these artifacts together with selected G-code evidence and produces the final anomaly label.}
\label{fig:framework}
\end{figure}

The framework is organized as three LLM-driven stages with a deterministic computational layer between them, as shown in Figure~\ref{fig:framework}. Extractor--LLM converts the G-code into a structured JSON object under a fixed process-parameter schema. Reference--LLM converts documentation for the intended printer and material into a compatible JSON object of recommended operating ranges. A deterministic comparison layer then aligns these two objects and constructs a deviation table. Judge--LLM receives the extracted parameters, reference ranges, deviation table, categorical flags, and the G-code itself; it returns the final class label together with an auditable rationale.

The decomposition has three methodological advantages. First, each stage has a narrow and testable interface: extraction and reference construction produce schema-constrained objects, while the comparison layer produces a numerical table. Second, role separation reduces the burden on the final judge; it no longer has to discover relevant parameters, infer reference ranges from memory, and perform all comparisons inside a single free-form response. Third, the intermediate artifacts are auditable. A user can inspect whether an error originated from parameter extraction, reference construction, numerical comparison, or final judgment, which is essential for manufacturing decision support.

\subsection{Extractor--LLM: structured parameter extraction}\label{sec:extractor}

Extractor--LLM reads a single G-code file and emits a JSON object containing process parameters relevant to material-extrusion quality. The schema covers thermal settings, extrusion-related quantities, layer heights, line-width channels, filament and nozzle geometry, retraction, travel behavior, fan policy, build-adhesion settings, and inferred printer/material identifiers. Explicit schema fields are used so that missing information appears as null values rather than as ambiguous narrative. In addition to predefined parameters, the model is prompted to report any additional parameter evidence or unusual G-code patterns as notes. Because G-code files can be very large and can exceed the context limits or practical cost limits of LLM calls, a deterministic shortening preprocessor preserves configuration blocks, initialization sequences, commands from the first three printed layers, and commands from the final printed layer.

After extraction, a normalization stage checks parameter units against the expected schema and converts them when needed. This step makes parameters extracted from different slicers and printers comparable with the reference ranges and reduces the risk that a downstream judgment is degraded by unit mismatches.

\subsection{Reference--LLM: manual-grounded reference extraction}\label{sec:reference}

Reference--LLM addresses a different source of knowledge, the operating envelope implied by printer and material documentation. Preliminary tests showed that recommended ranges inferred directly by general LLM backbones can differ from the ranges reported in the manuals for each printer--material combination. Manuals therefore provide an important grounding source for anomaly detection. Reference--LLM reads the documentation associated with a printer--material pair and provides recommended ranges under a schema aligned with the extractor output. Fields that are absent from the manuals remain null. In addition to prespecified parameters, Reference--LLM is prompted to report additional parameters, defect-related tips, warnings, factors, and parameter-interaction notes that can be extracted from the manuals. This stage makes the framework extensible because changing the printer, material, or AM process requires changing the documentation source rather than redesigning the judge.

\subsection{Judge--LLM: evidence-based anomaly judgment}\label{sec:judge}

Judge--LLM performs the final evidence integration. It consumes the extractor JSON, the reference JSON, the deterministic deviation table, categorical flags, and the shortened version of the G-code. Its role is not to rediscover all parameter values or reference ranges; its role is to interpret whether the structured evidence supports \texttt{ND}, \texttt{UE}, \texttt{OE}, \texttt{WP}, or \texttt{ST}.

The deviation table is one of the central structural artifacts of the framework. For each comparable parameter $k$, let $v_k$ denote the extracted value and $[\ell_k, u_k]$ the recommended range from Reference--LLM. The deviation table stores, for every $k$ that has both a numeric $v_k$ and a finite range:
\begin{equation*}
\Delta_k =
\begin{cases}
\ell_k - v_k, & v_k < \ell_k \\
0, & v_k \in [\ell_k, u_k] \\
v_k - u_k, & v_k > u_k,
\end{cases}
\qquad
\tilde{\Delta}_k = \frac{\Delta_k}{u_k - \ell_k}\ \text{when}\ u_k > \ell_k,
\end{equation*}
together with the raw $v_k$, $\ell_k$, $u_k$, the band width $u_k - \ell_k$, and a direction tag from \{below, in range, above\}. The table contains measurements and margins rather than preassigned defect labels. This design keeps quantitative comparison deterministic while leaving contextual interpretation to the final judge.

The Judge--LLM prompt uses a fixed output schema for the predicted label, confidence, support summary, deviations used, rationale, and evidence citations. This schema makes the final decision inspectable and allows downstream evaluation to distinguish a wrong class label from an incomplete rationale or missing evidence citation.



%% file: sections/implementation.tex
\section{Experimental Setup}\label{sec:experiment}

\subsection{Model slate}

The screening study uses three representative closed-source frontier backbones: GPT-5.4-mini (OpenAI)~\cite{openai2026gpt54mini}, Claude-Sonnet-4.6 (Anthropic)~\cite{anthropic2026claude46}, and Gemini-2.5-Flash (Google)~\cite{google2026gemini25flash}. These models were selected because they support long-context inputs and structured outputs, two requirements for G-code and documentation reasoning, while remaining in a comparable and affordable cost range. The sampling temperature is set to zero for all stages. The purpose of the model slate is not to rank vendors in general, but to test whether the proposed decomposition remains useful when the backbone assigned to each stage changes and to identify the strongest model for each framework role.

\subsection{Evaluation metrics}

All prompts, schemas, preprocessing rules, and evaluation samples are held fixed across provider configurations. Extractor--LLM is first evaluated independently on the $N=20$ extraction benchmark using tolerance-based accuracy and mean absolute percentage error (MAPE) over manually recorded ground-truth fields. The reference stage is evaluated through the downstream detection task because preliminary inspection showed that all candidate LLMs can extract numeric manual ranges reasonably well, whereas the usefulness of extracted notes and interaction guidance is best assessed through final anomaly-detection performance. Therefore, the strongest extraction configuration is used for the end-to-end screening study, while Reference--LLM and Judge--LLM are varied in a $3\times3$ grid. This design isolates the effect of reference construction and final judgment, which are the two stages most directly tied to end-to-end classification. The end-to-end metrics include accuracy, macro-F1, confusion matrices, per-class recall, and stratified accuracy by printer, material, and printer--material combination.

\subsection{Baselines and full-corpus experiment}

The primary baseline is a single-LLM classifier that receives only the shortened G-code and directly predicts the anomaly class without the extractor, reference, or deviation-table stages. Two prompt families are evaluated for this baseline on the $N=40$ screening set. The first follows the direct G-code classification style used in FDM-Bench~\cite{eslaminia2025fdmbench}. The second uses a stronger engineered prompt that asks the model to extract runtime-relevant G-code signals, compare support across all five classes, and select the best-supported label. The screening study is used to select configurations, and the full-corpus $N=200$ experiment is the final performance evaluation reported in this paper. This staged protocol separates model selection from final performance estimation and avoids using the full corpus for repeated configuration search, which can be expensive.

%% file: sections/case_studies.tex
\section{Results}\label{sec:case}

This section details the screening evaluation used to identify the optimal configuration and compares it with baseline models. Section~\ref{sec:results_extractor} evaluates the extraction role, Section~\ref{sec:results_grid} reports the Reference--LLM$\times$Judge--LLM configuration grid, Section~\ref{sec:results_cm} analyzes class- and domain-level behavior, Section~\ref{sec:results_baseline} compares against single-LLM baselines, and Section~\ref{sec:planned_full_corpus} reports the final full-corpus evaluation.

\subsection{Extractor--LLM: parameter-extraction benchmark}\label{sec:results_extractor}

Table~\ref{tab:llm1_summary} reports tolerance-based accuracy and MAPE for the three candidate extractors on the $N=20$ extraction benchmark subset. Tolerance-based accuracy counts a numeric extraction as correct when it falls within a predefined engineering tolerance around the manually recorded value; MAPE measures the magnitude of the remaining numerical deviation. The purpose of this experiment is to test whether current LLMs can reliably convert G-code into the structured representation required by the downstream framework. In addition to the full schema, a critical-parameter subset is reported because errors in extrusion, thermal, geometric, retraction, cooling, and adhesion fields have the greatest impact on the subsequent judgment.

\begin{table}[h]
\centering
\small
\setlength{\tabcolsep}{4pt}
\caption{Parameter-extraction benchmark for Extractor--LLM ($N{=}20$ G-codes, $18$ parameters per sample). All three backbones return a numeric prediction for every cell.}
\label{tab:llm1_summary}
\begin{tabularx}{\textwidth}{@{}>{\raggedright\arraybackslash}p{2.6cm}*{4}{>{\centering\arraybackslash}X}@{}}
\toprule
Backbone & Overall acc. & Critical acc. & Overall MAPE & Critical MAPE \\
\midrule
GPT-5.4-mini (OpenAI) & 0.944 & 0.939 & 3.50\% & 4.85\% \\
Claude Sonnet 4.6 (Anthropic) & 0.933 & 0.923 & 4.61\% & 6.39\% \\
Gemini 2.5 Flash (Google) & 0.911 & 0.900 & 6.05\% & 8.17\% \\
\bottomrule
\end{tabularx}
\end{table}

\begin{figure}[h]
\centering
\includegraphics[width=\linewidth]{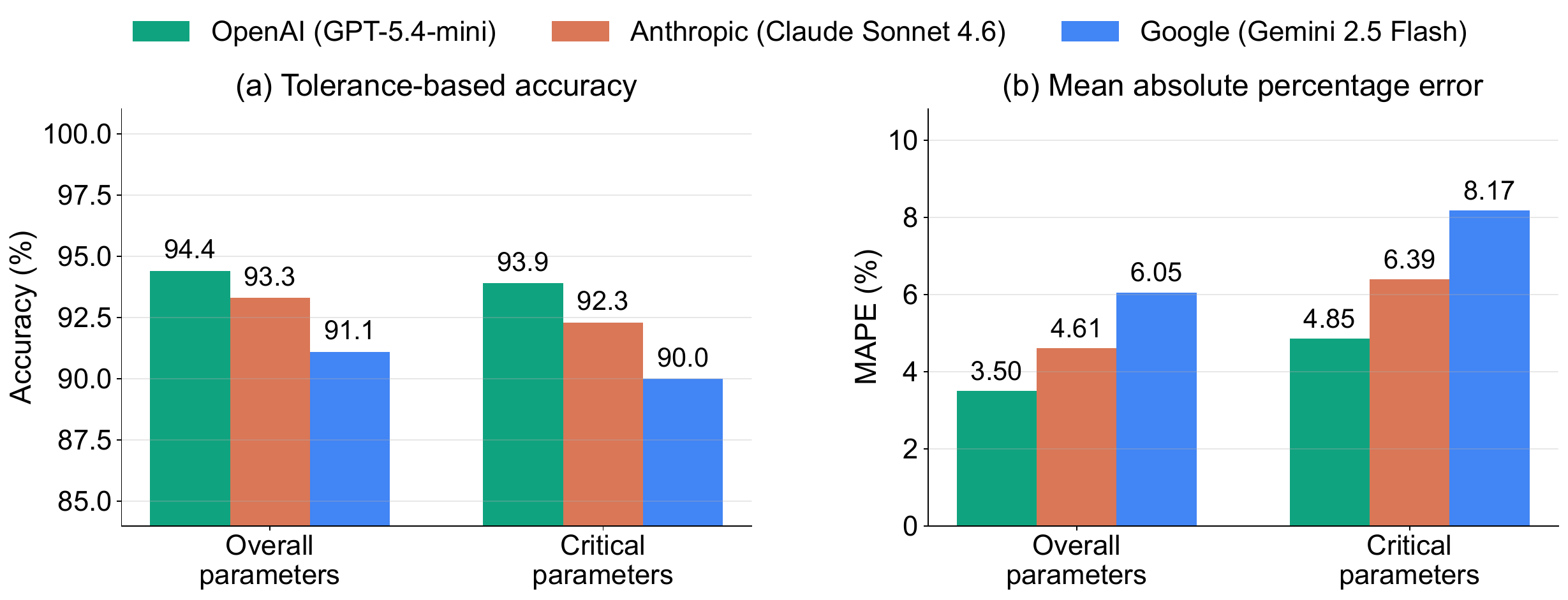}
\caption{Extractor--LLM benchmark with grouped comparisons at each position: (a) tolerance-based accuracy and (b) mean absolute percentage error for overall vs.\ critical parameter subsets. Within each group, bars correspond to the three provider backbones (legend), which isolates backbone differences at fixed subset definitions.}
\label{fig:llm1_summary}
\end{figure}

All three backbones exceed $90\%$ tolerance-based accuracy, indicating that structured parameter extraction is feasible with current frontier LLMs when the output schema is explicit. The accuracy differences are moderate, but the MAPE values show a clearer separation. GPT-5.4-mini has the lowest overall MAPE ($3.50\%$) and the lowest critical-parameter MAPE ($4.85\%$) compared with Claude Sonnet 4.6 and Gemini 2.5 Flash. For this reason, GPT-5.4-mini is selected as the Extractor--LLM backbone for the downstream Reference--LLM$\times$Judge--LLM screening study.

\subsection{End-to-end grid study (Reference--LLM and Judge--LLM)}\label{sec:results_grid}

Table~\ref{tab:grid_accuracy} reports end-to-end classification accuracy on the $N=40$ screening set across all nine Reference--LLM and Judge--LLM combinations. This grid treats model assignment as an experimental factor: the data, prompts, schemas, and deterministic comparison layer remain fixed, while only the backbones assigned to the reference and judgment roles vary.

\begin{table}[h]
\centering
\caption{End-to-end classification accuracy (\%) on the $N=40$ evaluation set across the $3\times 3$ Reference--LLM$\times$Judge--LLM grid, augmented with row-wise and column-wise averages. The best configuration pairs Gemini on both stages ($87.5\%$). Rows index Reference--LLM; columns index Judge--LLM.}
\label{tab:grid_accuracy}
\begin{tabular}{@{}lcccc@{}}
\toprule
Reference--LLM $\downarrow$~/~Judge--LLM $\rightarrow$ & OpenAI & Claude & Gemini & Row avg. \\
\midrule
OpenAI & 75.0 & 80.0 & 85.0 & 80.0 \\
Claude & 77.5 & 82.5 & 85.0 & 81.7 \\
Gemini & 65.0 & 85.0 & 87.5 & 79.2 \\
\midrule
Column avg. & 72.5 & 82.5 & 85.8 & 80.3 \\
\bottomrule
\end{tabular}
\end{table}

The grid supports two conclusions. First, the judge role dominates the end-to-end outcome. The Gemini judge has the highest column average ($85.8\%$), followed by Claude ($82.5\%$), while OpenAI as judge has the lowest column average ($72.5\%$). This pattern is consistent with the architecture: the judge is the only stage that must integrate extracted parameters, reference ranges, deviation magnitudes, and raw G-code evidence into a class-level decision. Second, reference construction is less variable than judgment once the reference output is mapped into the common schema. The row averages span a narrow range ($79.2$--$81.7\%$), indicating that the documentation-to-schema step is important but not the main source of performance differences in this screening set.

\subsection{Confusion matrices of leading configurations}\label{sec:results_cm}

\begin{figure*}[h]
\centering
\makebox[\textwidth][c]{\includegraphics[width=1.1\textwidth]{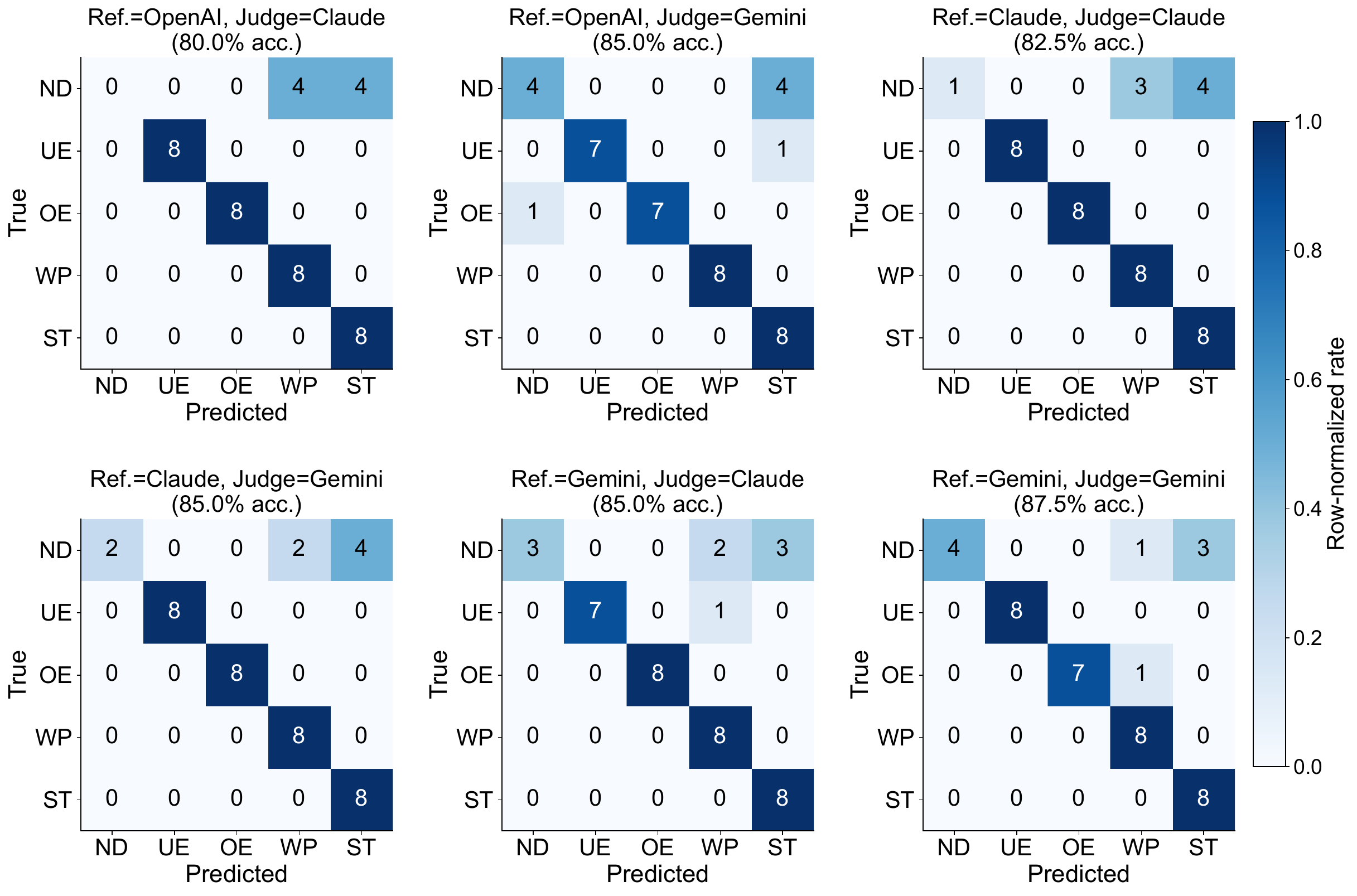}}
\caption{Confusion matrices ($5\times 5$) for the six leading Reference--LLM$\times$Judge--LLM configurations on the $N=40$ end-to-end set. Cell values denote raw sample counts; color encodes row-normalized rate. Each row corresponds to a true class with $8$ samples by construction.}
\label{fig:confusion_matrices}
\end{figure*}

\begin{figure}[h]
\centering
\makebox[\textwidth][c]{\includegraphics[width=1.05\textwidth]{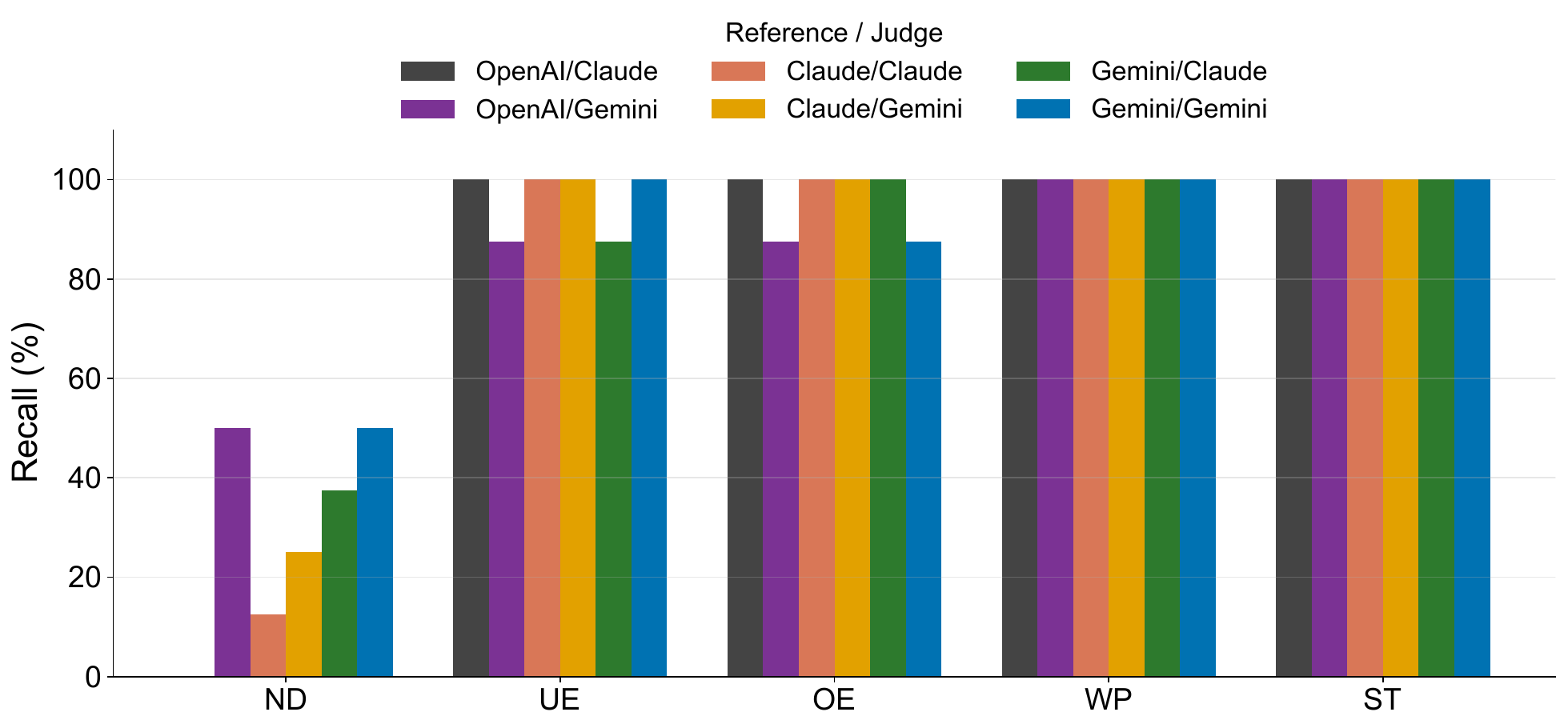}}
\caption{Per-class recall across the six leading Reference--LLM$\times$Judge--LLM configurations. Recall on UE, OE, WP, and ST is at or near ceiling for all configurations; the differentiator is the ND class.}
\label{fig:per_class_recall}
\end{figure}

Figure~\ref{fig:confusion_matrices} reports confusion matrices for the six leading configurations (all configurations with Judge--LLM $\in\{\text{Claude}, \text{Gemini}\}$). Figure~\ref{fig:per_class_recall} summarizes per-class recall across these configurations.

The figures support three findings. First, defect recognition is strong across the leading configurations. Recall on \texttt{UE}, \texttt{OE}, \texttt{WP}, and \texttt{ST} is at or near ceiling for most configurations, showing that the structured evidence produced by the framework is sufficient for the judge to identify the main defect mechanisms on this screening set.

Second, the non-defective class is the dominant source of residual error. The main failure mode is not confusion among defect types; it is conservative over-detection, where true \texttt{ND} samples are classified as \texttt{WP} or \texttt{ST}. This is visible in both the first row of Figure~\ref{fig:confusion_matrices} and the \texttt{ND} group in Figure~\ref{fig:per_class_recall}. The best \texttt{ND} recall is $50\%$ ($4/8$), achieved by (Gemini, Gemini) and (OpenAI, Gemini), while the worst is $0\%$ for (OpenAI, Claude). The framework is more likely to flag a benign file for review than to miss a defect, which is often preferable in pre-print screening but still leaves room for improving benign-case calibration.

\begin{figure}[h]
\centering
\makebox[\textwidth][c]{\includegraphics[width=1.05\textwidth]{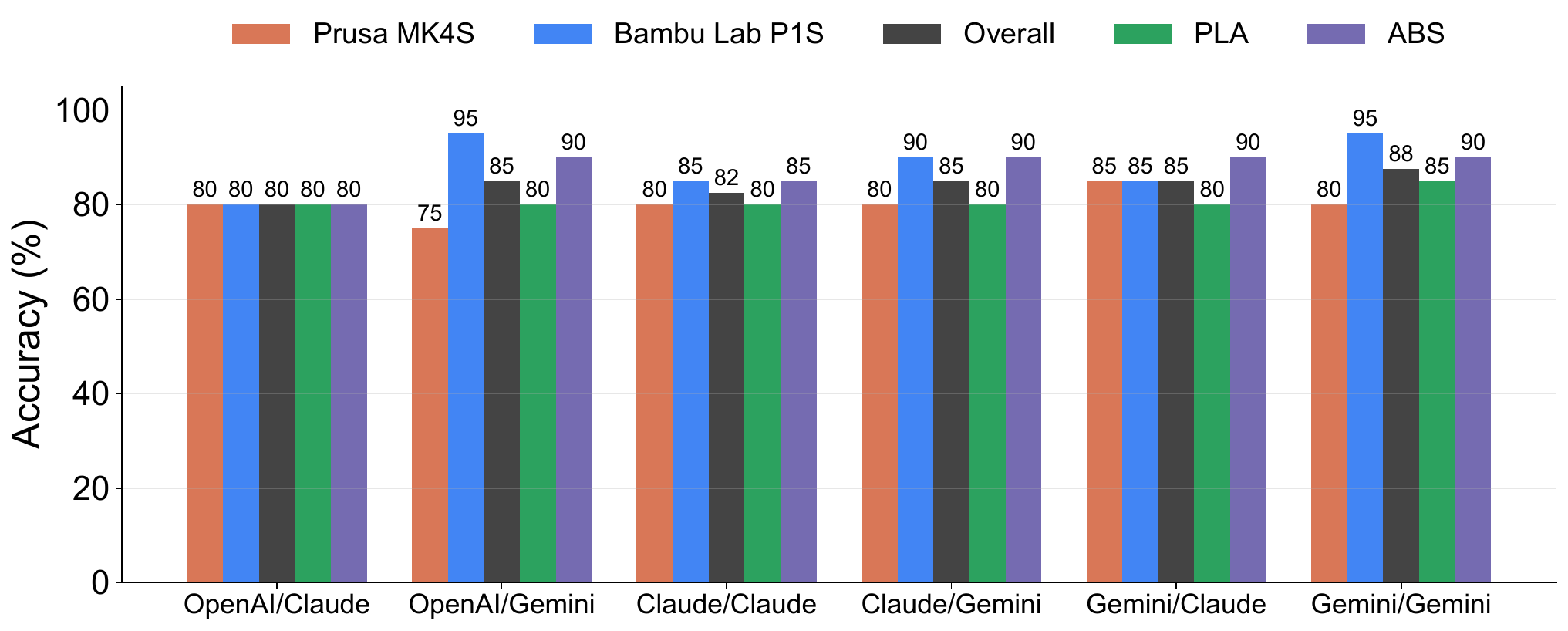}}
\caption{Integrated accuracy splits across the six leading Reference--LLM$\times$Judge--LLM configurations on the $N=40$ end-to-end set. For each configuration, bars report MK4S, BMP1, overall, PLA, and ABS. BMP1 is generally easier than MK4S, and ABS is generally easier than PLA across the leading runs.}
\label{fig:per_printer}
\end{figure}

Third, performance differences are structured by printer and material, not randomly distributed. Figure~\ref{fig:per_printer} reports printer, material, and overall accuracy in a single grouped chart. BMP1 tends to be easier than MK4S, and ABS tends to be equal to or easier than PLA. The best configuration (Gemini, Gemini) reaches $95\%$ on BMP1 and $80\%$ on MK4S, with $90\%$ on ABS and $85\%$ on PLA. These differences suggest that slicer conventions and material-specific parameter envelopes affect the clarity of the evidence available to the framework.

\begin{figure}[h]
\centering
\makebox[\textwidth][c]{\includegraphics[width=1.05\textwidth]{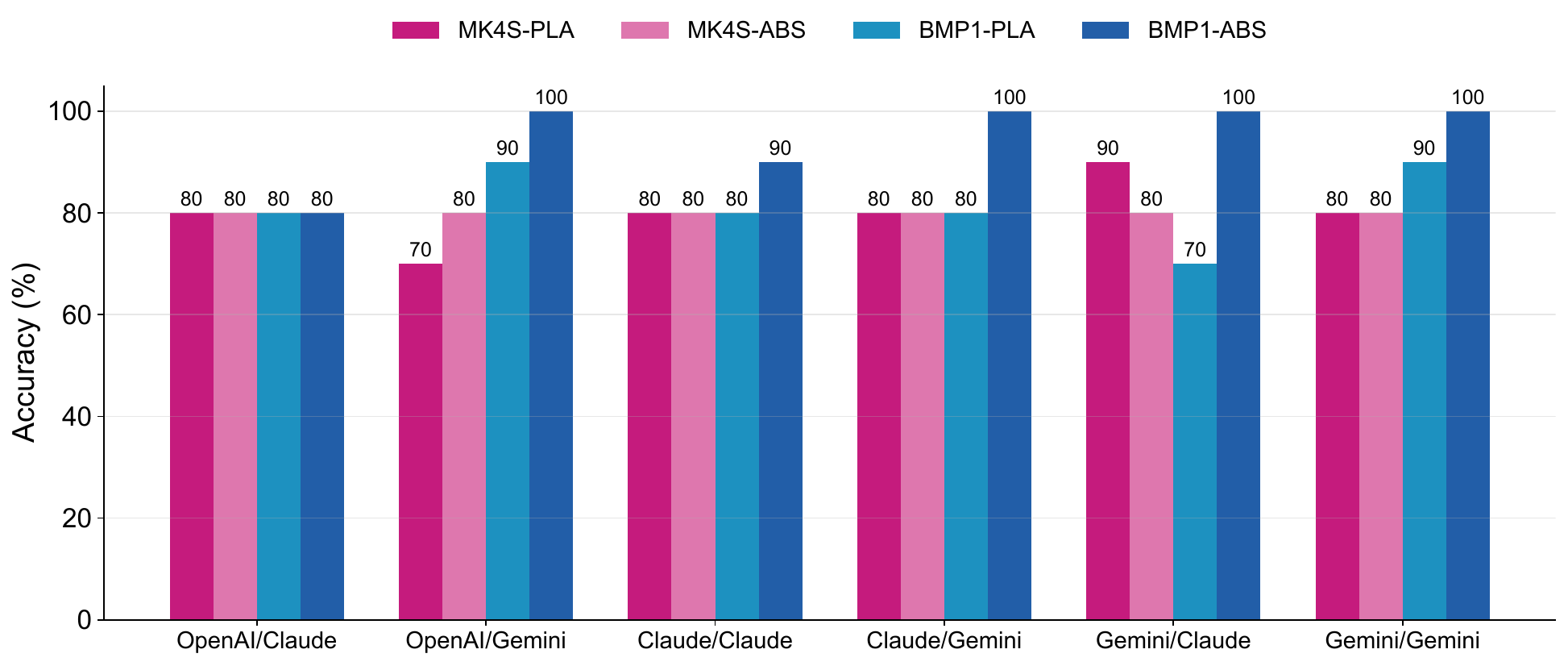}}
\caption{Accuracy by printer--material combination across the six leading Reference--LLM$\times$Judge--LLM configurations: MK4S-PLA, MK4S-ABS, BMP1-PLA, and BMP1-ABS. The largest variability appears on MK4S-PLA, while BMP1-ABS is near-ceiling for multiple configurations.}
\label{fig:per_combo}
\end{figure}

Figure~\ref{fig:per_combo} further resolves this interaction by reporting all four printer--material combinations. The dominant pattern is that BMP1-ABS remains the easiest slice, reaching $100\%$ in several configurations. In contrast, MK4S-PLA is more variable and often limits overall performance. This stratified view is important because an aggregate score can hide whether a configuration is robust across operating regimes or only strong on a favorable subset.

\subsection{Comparison with single-model baselines}\label{sec:results_baseline}

To quantify the value of decomposition, we evaluated each backbone as a single-model classifier that receives only the shortened G-code and directly predicts the final class label, without the extractor, reference, or deterministic comparison stages. Two prompt families are considered. The first follows the direct G-code classification style used in FDM-Bench~\cite{eslaminia2025fdmbench}. The second is an engineered single-model prompt that asks the LLM to extract runtime-relevant G-code signals, compare all five classes, and choose the most appropriate label. 

\begin{figure}[h]
\centering
\makebox[\textwidth][c]{\includegraphics[width=1.05\textwidth]{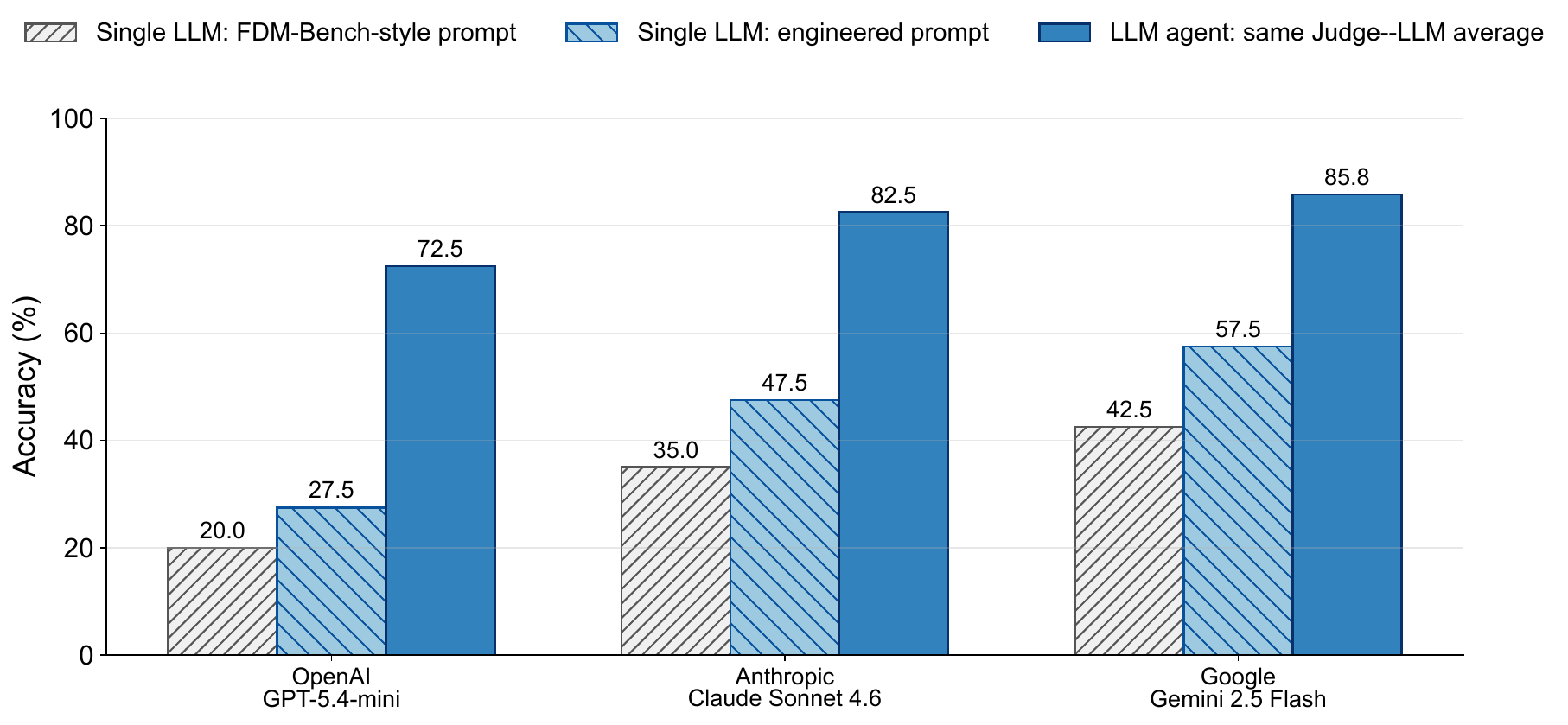}}
\caption{Comparison on the $N=40$ screening set between two single-LLM prompt families and the average framework accuracy obtained when the same provider is used as Judge--LLM across the three Reference--LLM choices. The single-LLM baselines receive only shortened G-code; the framework also uses extracted parameters, manual-grounded references, and deterministic deviation tables.}
\label{fig:multi_vs_single}
\end{figure}

Figure~\ref{fig:multi_vs_single} shows that direct prompting is a weak baseline for this task. With the FDM-Bench-style prompt, GPT-5.4-mini, Claude-Sonnet-4.6, and Gemini-2.5-Flash obtain $20.0\%$, $35.0\%$, and $42.5\%$ accuracy, respectively, on the $N=40$ screening set. Prompt engineering improves the single-model results to $27.5\%$, $47.5\%$, and $57.5\%$, with Gemini again the strongest baseline. However, the framework remains substantially higher: averaging over the three Reference--LLM choices gives $72.5\%$ for OpenAI as Judge--LLM, $82.5\%$ for Claude, and $85.8\%$ for Gemini. The comparison shows that the performance gain is not merely a result of writing a stronger direct prompt; it comes from decomposing extraction, reference grounding, numerical comparison, and judgment into separate auditable steps.

\subsection{Full-corpus evaluation}\label{sec:planned_full_corpus}

The $N=40$ study is used for model selection and error analysis, and the selected configurations are then evaluated on the full $N=200$ corpus. Three framework configurations are included in this final comparison. Gemini--Gemini is selected because it has the highest screening accuracy and the strongest Judge--LLM column average. Claude--Gemini is selected because it is the second-best Gemini-judge configuration and maintains stable performance across printer--material combinations. Claude--Claude is included as a third framework configuration due to its high performance with no printer--material combination accuracy below $80\%$ in the screening set. Gemini--Claude is not selected despite its competitive screening accuracy because one printer--material combination shows substantially lower accuracy, which is less desirable for deployment. For the single-LLM baseline, the engineered Gemini prompt is selected because it is the best direct classifier in the $N=40$ baseline screen.

\begin{figure}[h]
\centering
\includegraphics[width=0.95\linewidth]{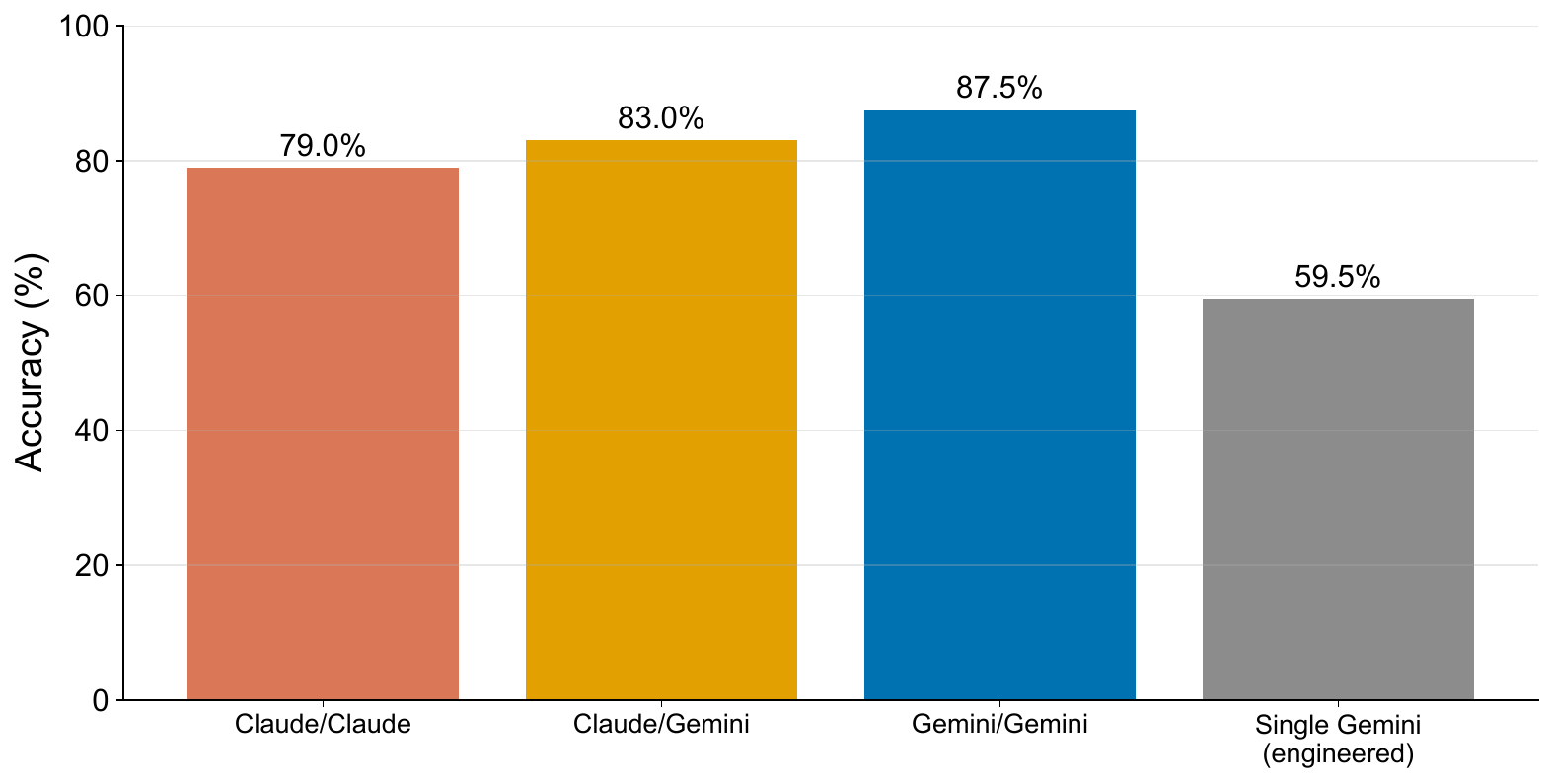}
\caption{Full-corpus overall accuracy for the selected framework configurations and the selected engineered Gemini single-LLM baseline on $N=200$ G-codes.}
\label{fig:full_corpus_accuracy}
\end{figure}

\begin{figure*}[h]
\centering
\makebox[\textwidth][c]{\includegraphics[width=0.9\textwidth]{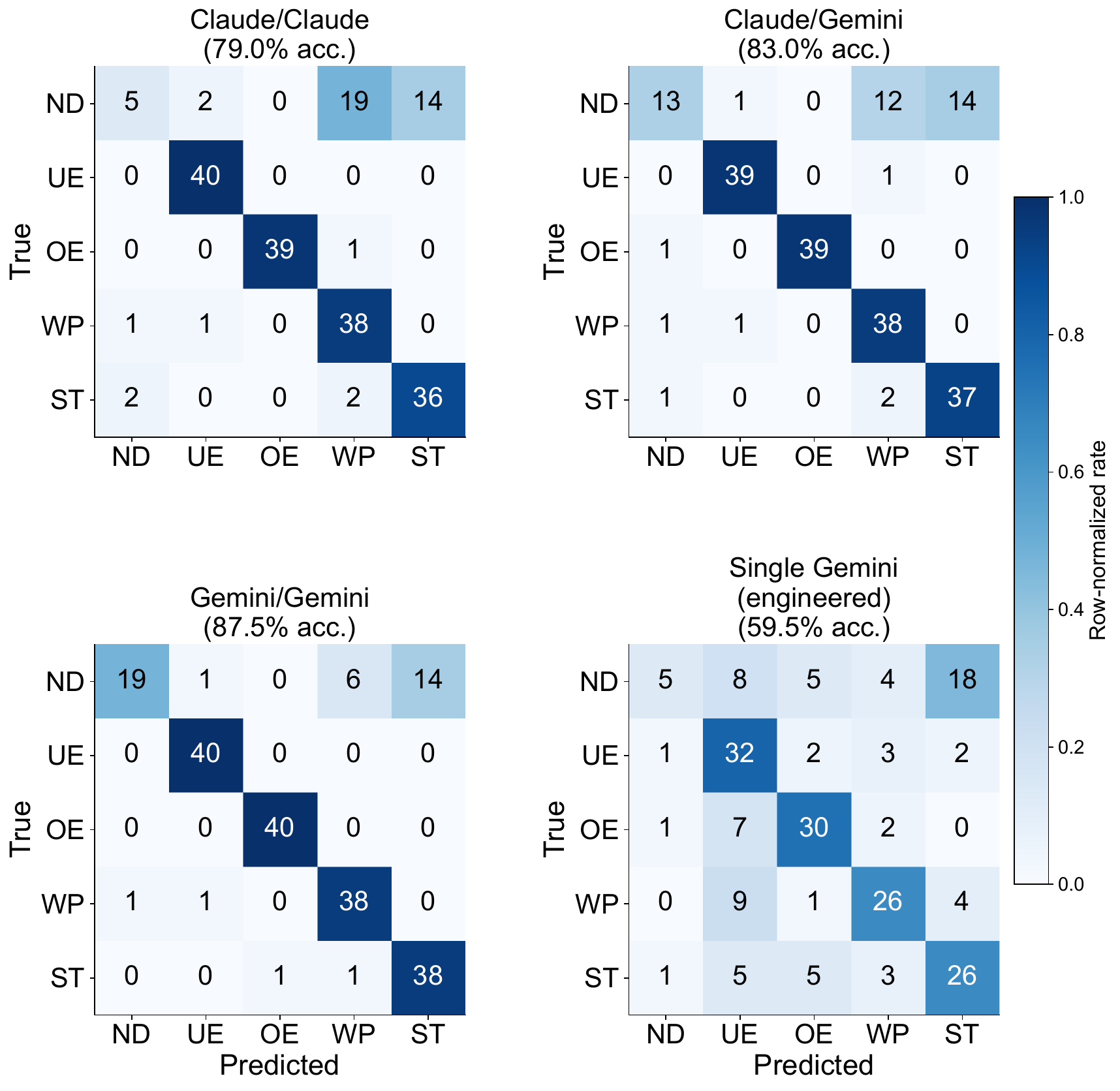}}
\caption{Full-corpus confusion matrices for the selected framework configurations and the engineered Gemini single-LLM baseline. Gemini--Gemini has the highest overall accuracy and the strongest ND recall among the framework candidates, while the single-LLM baseline shows broader confusion across defect classes.}
\label{fig:full_corpus_confusion}
\end{figure*}

\begin{figure}[h]
\centering
\includegraphics[width=\linewidth]{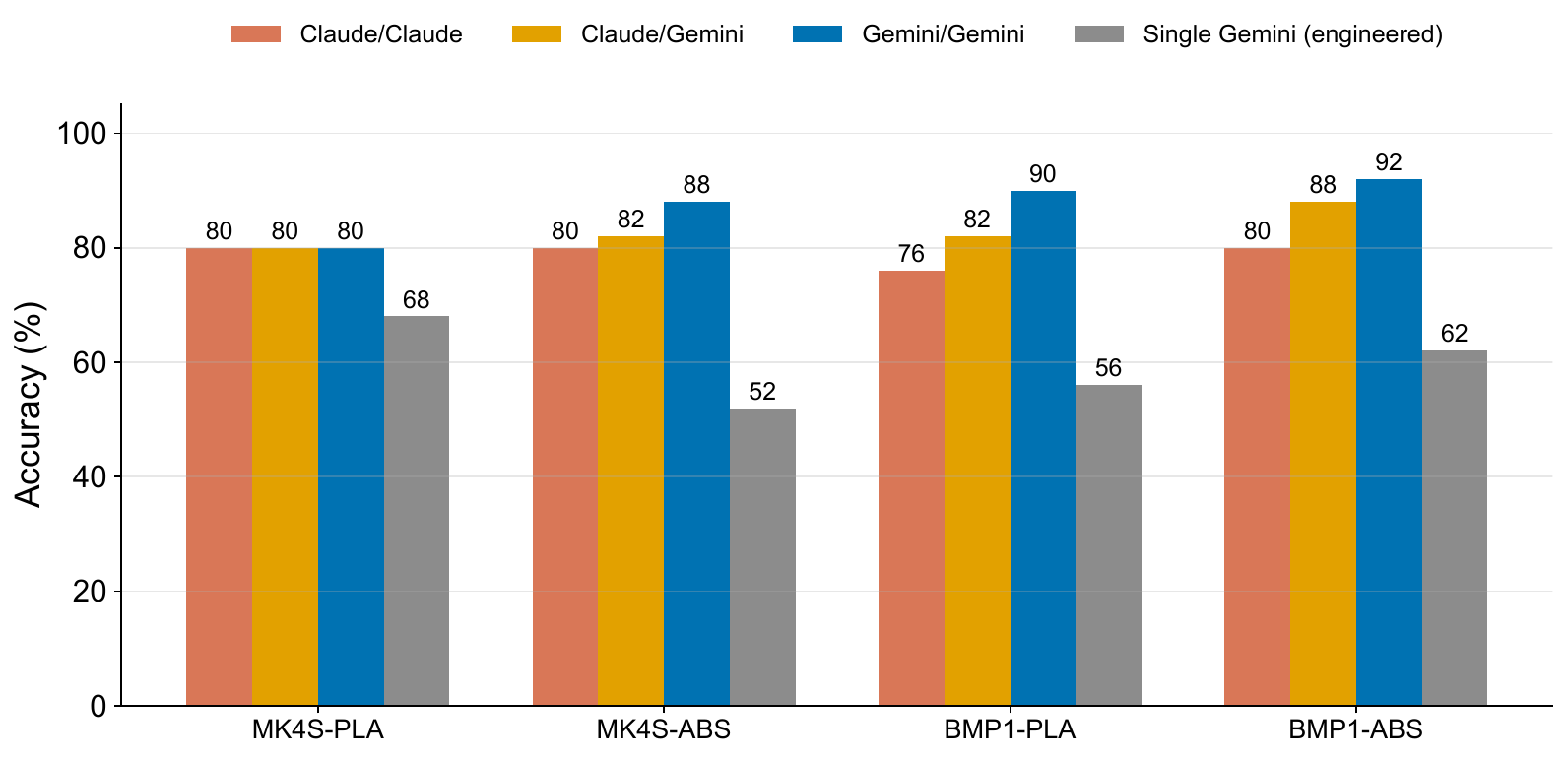}
\caption{Full-corpus accuracy by printer--material combination for the selected framework configurations and the engineered Gemini single-LLM baseline. Gemini--Gemini is the strongest configuration on BMP1-PLA and BMP1-ABS and improves MK4S-ABS relative to the other candidates.}
\label{fig:full_corpus_combo}
\end{figure}

The full-corpus results confirm the ranking suggested by the screening study. Gemini--Gemini achieves the highest overall accuracy at $87.5\%$, followed by Claude--Gemini at $83.0\%$ and Claude--Claude at $79.0\%$. The selected engineered Gemini single-LLM baseline reaches $59.5\%$ accuracy on the same $N=200$ corpus. The best framework therefore improves overall accuracy by $28.0$ percentage points over the strongest direct baseline while using the same shortened G-code evidence plus the agent's structured extraction, reference, and comparison artifacts. Figure~\ref{fig:full_corpus_combo} illustrates that the improvement is visible across all printer--material combinations. Gemini--Gemini reaches $80\%$, $88\%$, $90\%$, and $92\%$ on MK4S-PLA, MK4S-ABS, BMP1-PLA, and BMP1-ABS, respectively, compared with $68\%$, $52\%$, $56\%$, and $62\%$ for the engineered Gemini baseline. The main remaining framework challenge is ND recall, as shown in Figure~\ref{fig:full_corpus_confusion}; nevertheless, Gemini--Gemini improves ND detection relative to the two Claude-reference candidates and the single Gemini model, while preserving near-ceiling recall for defect classes.

%% file: sections/discussion.tex
\section{Discussion}\label{sec:discussion}

\subsection{Interpreting the screening results}

The screening and full-corpus results show that the proposed decomposition is most effective when the anomaly evidence is directional and class-specific. UE, OE, WP, and ST are detected reliably across the leading configurations because their parameter deviations form coherent patterns in the deviation table and supporting G-code evidence. The remaining difficulty is the ND class. A benign file may contain small deviations from nominal settings, and the judge must decide whether those deviations are harmless or early evidence of a defect. The leading configurations are conservative, preferring to flag some borderline-benign samples as WP or ST rather than incorrectly accept them as ND. For pre-print screening, this is a defensible operating point because a false alarm leads to review, while a missed defect can waste material and machine time. Nevertheless, improving benign-case calibration is the most important methodological target before deployment.

\subsection{Why decomposition matters}

The large gap between the multi-stage framework and the single-model baselines should be interpreted as evidence for task decomposition rather than as evidence for a particular vendor model or prompt template. Even after prompt engineering, the single-model baseline must infer parameters, estimate reference ranges, compare values, and classify the file in one step using only shortened G-code. The proposed framework externalizes these operations. Parameters are extracted into a schema, documentation is translated into aligned reference ranges, and numerical margins are computed deterministically before judgment. This turns the final LLM call into an evidence-integration task rather than an unconstrained code-reading task. The result is not only higher accuracy, but also a more inspectable failure mode because errors can be traced to extraction, reference interpretation, comparison, or final classification.

\subsection{Generalization and deployment}

Although the experiments use two printers and two materials, the method is not restricted to these machines. The core requirement is that the manufacturing process have a textual program or recipe, process parameters that can be extracted into a schema, and reference knowledge that can be converted into operating ranges. FFF is a suitable first domain because G-code is explicit and because desktop FFF has a large non-expert user base. The same design logic could extend to other AM settings where textual build files and reference envelopes are available, but the extraction schema and reference documents would need to be adapted and validated.

The staged architecture also has practical advantages. Reference construction is performed per printer--material pair rather than per file, and extraction artifacts can be retained for audit and reproducibility. At scale, the final judgment call is the dominant per-file cost. 

\subsection{Limitations and future work}

The first limitation is that the framework depends on access to relevant printer and material documentation. This manual-grounded design improves the reliability of reference ranges, but it reduces the autonomy of the system because the appropriate manuals must be identified and provided before inference. A natural extension is to automate the documentation-search step so that the system can search, retrieve, verify, and version the relevant reference sources with minimal operator input.

The second limitation is that the current framework detects and explains likely G-code anomalies but does not yet repair them. After the system assigns a defect class, an operator still needs to modify the slicer settings or G-code to correct the issue. A direct future extension is an automated correction stage that proposes safe parameter changes or edited G-code after the anomaly type and supporting evidence have been identified.

%% file: sections/conclusion.tex
\section{Conclusion}\label{sec:conclusion}

This paper presented a generalizable LLM agent framework for pre-print anomaly detection in AM. The framework treats machine program screening as a structured reasoning problem rather than a single text-classification prompt. Extractor--LLM maps the machine program to a process-parameter schema, Reference--LLM maps documentation to aligned operating ranges, and Judge--LLM interprets deterministic deviation evidence to assign an anomaly label. This decomposition is intended to be general across printer and material settings because the task-specific knowledge is carried through schemas, reference documents, and intermediate comparison artifacts rather than embedded in one monolithic prompt.

Evaluation on the $N=200$ FFF G-code corpus shows that Gemini--Gemini is the strongest framework configuration with $87.5\%$ accuracy, followed by Claude--Gemini at $83.0\%$ and Claude--Claude at $79.0\%$, while the selected engineered Gemini single-LLM baseline reaches $59.5\%$. Thus, the best framework improves the final full-corpus accuracy by $28.0$ percentage points over the strongest direct-prompt baseline. The results indicate that the judgment role is the most influential stage, that defect classes are detected reliably in leading configurations, and that remaining errors are concentrated in conservative false alarms on ND samples. Beyond FFF, the central idea -- combining LLM-based interpretation of text programs and documentation with deterministic intermediate representations -- is applicable to other AM processes where machine instructions can be screened before execution.

%% file: sections/back_matter.tex
\section*{Acknowledgments}
Ahmadreza Eslaminia and Klara Nahrstedt acknowledge support from the U.S.\ National Science Foundation under Grants CNS-2106592, OAC-2126246 (MAINTLET), and CNS-2437204, and from subaward UCDavis A21-0845-S003.
Chenhui Shao acknowledges support from the U.S. National Science Foundation under Award No. 2434383.

\bibliography{mybibfile}